# Predicting the Financial Impact of Supply Chain Risk for Major AI-Related Semiconductor Firms: A Heterogeneous Graph Patch Transformer Approach

Jianna Hur and Sagar Samtani
Data Science and Artificial Intelligence Lab (DSAIL), Kelley School of Business
Indiana University, Bloomington, IN, USA

## Abstract

Modern semiconductor production relies on a globally distributed, multi-tier supply chain in which financial stress at one firm spreads with a delay and eventually affects the revenue, inventory, and profitability of the companies that design AI chips. Most firms see only their direct partners, and prior predictive research has mainly targeted market-based risk measures, so few tools forecast how supply chain stress will appear in reported financials. In this study, we propose a heterogeneous graph patch transformer that forecasts these quarterly changes one and two quarters ahead. Learning from a 15,186-company network over 60 quarters, the proposed model fuses quarterly fundamentals with macro-trade, event, and disaster signals through learned gates, carries risk across supplier, customer, ownership, and headquarters relations through typed, direction-specific propagation, and encodes the propagated histories with patch-based tokenization. In preliminary experiments on 116 focal semiconductor firms, the proposed model achieves the lowest error on every target at both horizons, and its profitability advantage widens at the two-quarter horizon. These forecasts can help supply chain managers and investors act before disruptions appear in reported financials.

## I. Introduction

Semiconductors are the foundation of AI and the broader digital economy. The market reached about $628 billion in 2024 and keeps growing as AI adoption accelerates [45]. Production depends on a globally distributed value chain in which fabless designers rely on foundries, equipment makers, and material suppliers across capital-intensive, multi-tier pipelines [37, 44]. Shocks in such networks do not remain local, and disruptions at individual firms can grow into substantial output losses for their customers and suppliers [1, 5].

Financial stress is one of the most consequential forms of this propagation because it travels in both directions of the value chain with a delay. The current AI cycle illustrates this pattern. Surging GPU demand first strained foundry and packaging capacity, and within two to three years the constraint moved to memory, where prices were expected to rise by another 50 percent in a single quarter [47]. The pressure is now reaching AI infrastructure, where data center demand is projected to nearly triple by 2030 and lead times for critical grid equipment stretch to multiple years [3, 39, 43]. A firm that can anticipate where financial stress will emerge next can adjust sourcing, inventory, and product portfolios in advance.

Most companies nonetheless have limited visibility into their own supply networks. In a recent global survey, the majority of companies understood their supply chain risks only up to tier one [38], even though advanced semiconductors are frequently single-sourced [4]. Such visibility carries measurable value, since firms that digitalized their supply chain management significantly reduced their expected default frequency [55]. Analysts fill only part of the gap, because the few who follow both a buyer and its supplier act as information intermediaries [9].

Academic research provides useful starting points, but extant studies have three key limitations for this problem. First, supply chain risk management (SCRM) research offers mature frameworks for identifying and assessing disruptions and now detects disruption signals at scale [10, 17, 23], but it rarely estimates the financial consequences for connected firms. Second, finance research shows that supply chain shocks move the stock returns of linked firms [12, 22, 52], and IS research predicts financial risk from rich multisource data [48, 54], yet both mainly target market-based measures or regulatory labels that are only indirectly tied to operating outcomes. Third, the networks in prior work are often small, single-typed, or generic, whereas the semiconductor setting is multi-tier, heterogeneous, and country-dependent at the same time.

In this study, we forecast the quarterly financial impact of supply chain risk across the semiconductor value chain, measured as changes in revenue, inventory, and profitability one and two quarters ahead. We propose a heterogeneous graph patch transformer that mirrors this propagation process in three components: (1) gated multi-source fusion, which encodes each firm's quarterly fundamentals together with macro-trade, event, and disaster signals, (2) typed, direction-specific propagation, which carries risk across supplier, customer, ownership, and headquarters relations with a separate transmission strength per relation, and (3) patch-based delta forecasting, which maps the propagated histories to quarterly changes while preserving multi-quarter patterns. Against machine learning, temporal, graph, and hybrid benchmarks on 116 focal semiconductor firms, the proposed model achieves the lowest error on every target at both horizons, reducing revenue mean absolute error (MAE) by 23.6 percent against the strongest baseline one quarter ahead, and its profitability advantage widens at two quarters, consistent with the delayed transmission that motivates a network model.

## II. Literature Review

We review three literatures, namely SCRM studies on detecting disruptions, financial risk studies in supply chain settings, and the graph and patch transformer techniques that form the methodological basis of our design.

### 2.1. Supply Chain Risk Identification and Assessment

SCRM research describes disruption risk through cycles of identification, assessment, mitigation, and monitoring across financial, supply, demand, and operational categories [6, 23]. Recent studies detect disruption signals computationally and at scale, from technical documents and news that reveal disruptions in EV battery supply chains [10] to earnings calls distilled into firm-level risk measures with generative language models [17], and the field's scope has deepened to sub-tier structure, where tier-two conditions causally affect buyer performance [14].

What detection does not yet deliver is magnitude. Studies that connect supply chain conditions to financial outcomes stop short of firm-level forecasts. Digitalizing supply chain management lowers a firm's own expected default frequency [55], and disasters produce output losses that propagate to connected firms [5], but both estimate average effects after the fact rather than predicting how large the financial response of a specific connected firm will be once a disruption is detected. Answering that question requires a clear definition of the financial outcomes to predict, so we review the financial risk literature next.

### 2.2. Financial Risk in Supply Chain Settings

Financial risk refers to a firm's vulnerability to adverse financial outcomes, such as distress, default, or loss of firm value, arising from its financial structure, profitability, and market conditions [48, 54]. Prior studies measure this vulnerability with accounting indicators, regulatory labels such as China's Special Treatment (ST) designation, and market-based measures such as volatility and crash risk [21, 34]. The IS community has expanded firm-level prediction along two axes. One axis enriches the signals, from textual disclosures for bankruptcy prediction [36] to multisource financial data [48] and vocal cues from earnings calls [54]. The other axis enriches the structure, from interfirm networks that predict labor market competition [33] to time-varying neighbor influence in financial risk prediction [49]. Along both axes, richer signals and network information consistently improve accuracy.

The finance literature establishes why network information should matter, and its findings share one structure, namely that shocks at one firm surface in the outcomes of its partners with a delay. Distress depresses the wealth of suppliers and customers [22], returns of linked firms adjust slowly enough to be predictable [12], systematic risk aggregates through network position [41], and output losses propagate along supply chains after disasters [5]. Even unobserved sub-tier structure shapes focal-firm equity risk [51, 52]. Firm-level shocks therefore do not diversify away, and a firm's network position determines its exposure.

Most of these studies, however, target market-based or regulator-defined outcomes. Volatility moves with market-wide sentiment, which obscures firm-specific supply chain effects, and the ST label applies only to Chinese listed firms. Neither captures the changes in revenue, inventory, or profitability through which a disruption reaches operations, and the delayed, structured propagation documented above is rarely represented inside the prediction model. We therefore adopt fundamental outcomes as targets and forecast them with a model that represents the network explicitly, which requires the methods we review next.

### 2.3. Graph Representation Learning and Patch Transformer Techniques

Graph representation learning encodes the entities and relationships of a network into vectors that preserve its structure [30, 57]. The prevailing family is the message-passing graph neural network (GNN), in which each layer aggregates neighbor representations, so two layers can carry a tier-two shock to the focal firm. Two properties limit standard GNNs in our setting. First, all neighbors share aggregation weights, so supplier and customer relations collapse into one semantic space even though shocks travel differently along each. Second, repeated local averaging causes over-smoothing, which weakens exactly the long-range influences that motivate a network model [29].

Graph transformers replace uniform local aggregation with attention, so influential partners receive more weight and distant dependencies survive stacking [42]. The heterogeneous graph transformer (HGT) further assigns separate projections to every node type and a separate attention matrix to every edge type, so a supplier edge and a customer edge learn different, direction-specific transmission strengths [25]. This typed, direction-specific propagation carries risk across supplier, customer, ownership, and headquarters relations without collapsing their meanings.

Capturing how the propagated signals evolve requires a temporal model. Recurrent models such as long short-term memory (LSTM) [24] and the gated recurrent unit (GRU) [11] compress the entire history into one gated memory state, which suits short-term persistence but degrades over longer ranges, a limitation documented in financial and traffic applications alike [18, 19]. Attention removes this bottleneck by letting any two time steps interact directly, and the Transformer stacks self-attention and feed-forward layers on this mechanism [46]. Applied to time series, however, the standard design treats every step as one token, which carries little meaning on its own and lets a heavily parameterized model overfit short histories. The patch time series transformer (PatchTST) instead groups $P$ consecutive steps into one token with stride $S$, which preserves the local shape of multi-period patterns and reduces the tokens entering attention from $T$ to roughly $T/S$ [40]. These properties suit financial panels with only 60 quarterly observations per firm.

Patching has since become the default tokenization, and its extensions relax the fixed window and couple patch encoders with graph structure for long-horizon forecasting and anomaly detection [13, 28, 35, 50, 53, 56]. The graphs in this line, however, remain homogeneous sensor or variable networks, and none targets firm-level financial outcomes over typed multi-tier business relationships. We therefore combine typed, direction-specific propagation with patch-based temporal encoding, two techniques that each literature developed in isolation. We formalize the resulting gaps next.

## III. Research Gaps and Questions

We identified three research gaps. First, SCRM research detects disruptions at scale but rarely embeds the detected signals in a model that forecasts focal-firm financial impact. Second, financial risk research relies on market-based or regulator-defined labels and seldom isolates multi-tier mechanisms. Third, graph-based methods capture heterogeneous structure or short-series temporal dynamics, but not both in one forecasting model. We therefore propose two research questions.

RQ1. How do proximity to shocked nodes and shared sub-tier dependencies shape a focal firm's financial response to exogenous shocks across a multi-tier semiconductor network?

RQ2. How can heterogeneous, multi-frequency signals from financial, macro-trade, event, and disaster sources be integrated within a graph-based framework to improve quarterly financial impact forecasts relative to firm-level baselines?

## IV. Proposed Research Design

Based on the identified research gaps, we propose a heterogeneous graph patch transformer framework for financial impact forecasting. Figure 1 summarizes the proposed research design, which consists of three components. The first collects multi-source data, aligns it to a quarterly grid, and constructs the heterogeneous network, the second is the proposed model, and the third evaluates it against benchmarks. We describe each in turn.

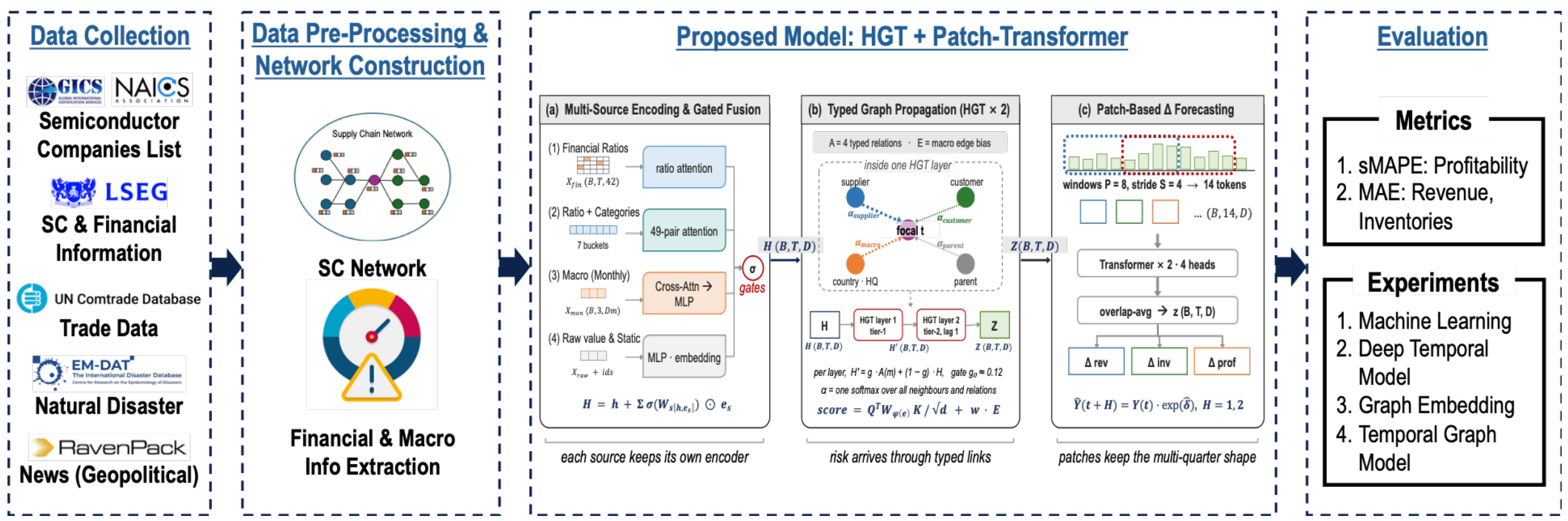

**Figure 1. Proposed Research Design Overview**

### 4.1. Data Collection

We collected 3,450 public companies whose GICS or NAICS classifications carry semiconductor keywords from LSEG Refinitiv and treated them as candidate focal firms. Their supplier, customer, and ownership relationships come from the LSEG Refinitiv Value Chains database, which reports counterparty identities with relationship evidence counts and confidence scores, and the network was constructed outward from these firms. Table 1 summarizes the data, whose frequencies range from daily to quarterly and motivate the alignment step in Section 4.2.

| Category | Data | Source | Information | Frequency |
|---|---|---|---|---|
| Macro Factors | Natural Disaster information | EM-DAT | Location, date, damage size, and other details for 9,323 natural disasters | Irregularly* |
| | Events for each country | RavenPack | Keyword-based pre-classified events | Daily |
| | Trade Data | UN Comtrade | Trade data for 5 countries (US, China, Taiwan, Korea, Japan) | Monthly |
| Company List | List of relevant companies with industry information | LSEG (Refinitiv) | 3,450 public companies with 'semiconductor' keywords in GICS or NAICS industry classifications | Irregularly* |
| Supply Chain (SC) | Suppliers-Customers relationship | LSEG (Refinitiv) | 27,926 unique directed relationships | Irregularly* |
| Financial Information | Each company's financial performance | LSEG (Refinitiv) | Financial statements (Income Statement, Balance Sheet, Cash Flow) | Quarterly |

**Table 1. Data Collection (*Irregularly: Updated with no fixed frequency)**

Table 1 reports 3,450 collected companies and 27,926 collected SC relationships. With 4,503 ownership and 1,296 headquarters links added, the network holds 15,186 companies and 89 countries connected by 33,725 typed edges, and 1,849 companies enter as focal firms with value-chain records. Each company carries headquarters country and primary industry attributes.

Firm fundamentals were collected quarterly from LSEG Refinitiv for every firm in the network. We computed 42 financial ratios organized in seven categories, namely profitability, development capacity, solvency, operational capabilities, capital expansion, finance structure, and trade credit. The first 39 follow Wang et al. [48], and the three added trade credit ratios that capture supply chain payment terms follow Ersahin et al. [15]. We retained raw revenue, inventory, and profitability values as prediction targets over a fixed grid of 60 quarters from 2010 to 2024.

Three external sources capture macro-level shocks. RavenPack provides daily country-level event signals with 47.2 million events between 2010 and 2024, EM-DAT contributes 8,600 of its 9,323 collected natural and industrial disaster records that fall inside the 2010 to 2024 window, and UN Comtrade provides monthly bilateral trade flows for 25 semiconductor-relevant Harmonized System (HS) codes selected per the guideline of Korea Customs Office [32].

### 4.2. Data Pre-Processing and Network Construction

All signals are aligned to the quarterly grid. Daily events and monthly trade flows are aggregated into quarterly country summaries, and each quarterly ratio is paired with an observation mask, so missing values are marked rather than imputed. Every series is standardized with statistics computed on the training window only, which later map forecasts back to original units. We then construct a heterogeneous network with firms and countries as nodes. Figure 2 summarizes the four relation types around the focal firms and the country nodes that carry the quarterly shock.

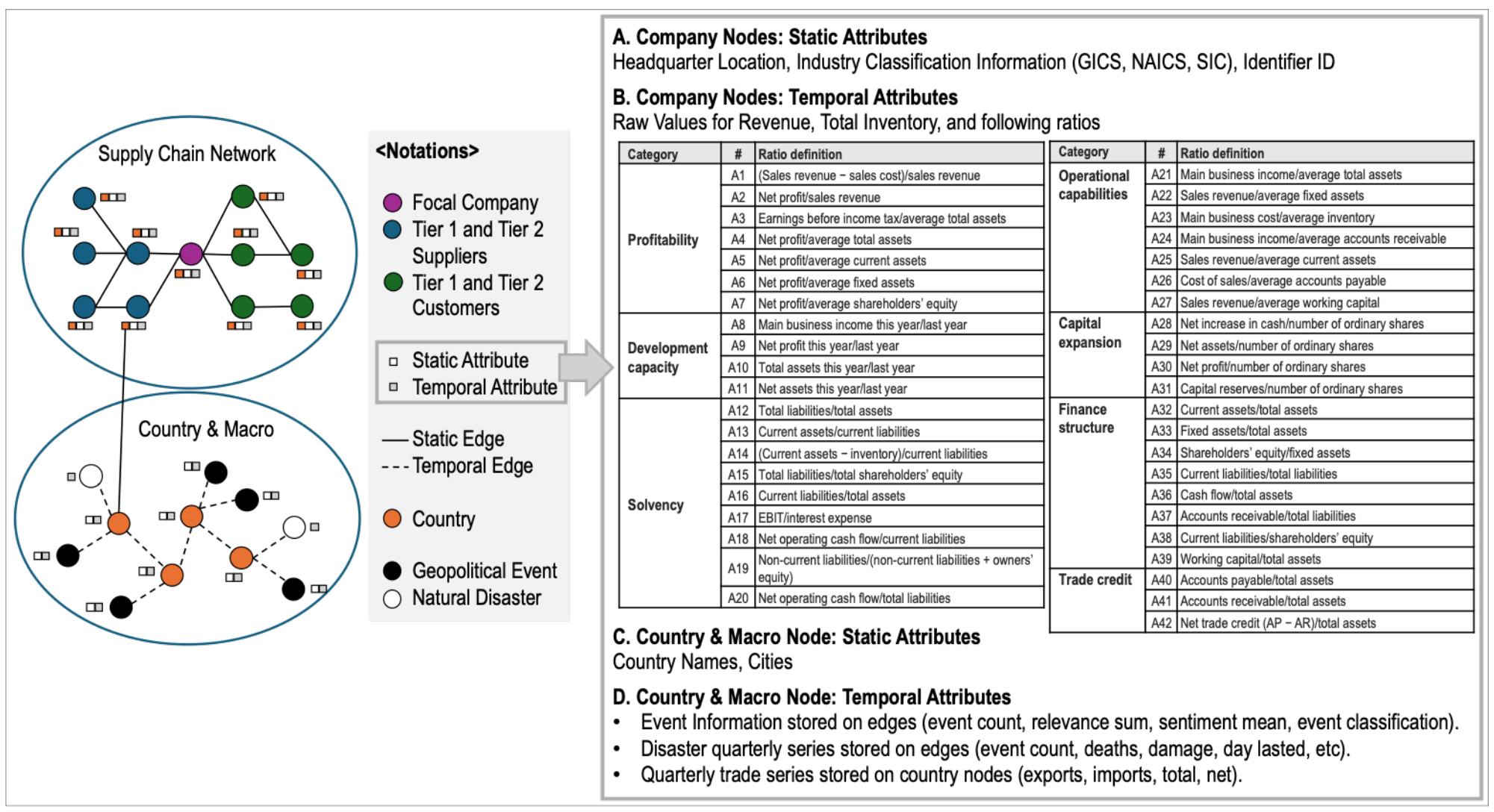

**A. Company Nodes: Static Attributes**
Headquarter Location, Industry Classification Information (GICS, NAICS, SIC), Identifier ID

**B. Company Nodes: Temporal Attributes**
Raw Values for Revenue, Total Inventory, and following ratios

| Category | # | Ratio definition |
|---|---|---|
| Profitability | A1 | (Sales revenue − sales cost)/sales revenue |
| | A2 | Net profit/sales revenue |
| | A3 | Earnings before income tax/average total assets |
| | A4 | Net profit/average total assets |
| | A5 | Net profit/average current assets |
| | A6 | Net profit/average fixed assets |
| | A7 | Net profit/average shareholders' equity |
| Development capacity | A8 | Main business income this year/last year |
| | A9 | Net profit this year/last year |
| | A10 | Total assets this year/last year |
| | A11 | Net assets this year/last year |
| Solvency | A12 | Total liabilities/total assets |
| | A13 | Current assets/current liabilities |
| | A14 | (Current assets − inventory)/current liabilities |
| | A15 | Total liabilities/total shareholders' equity |
| | A16 | Current liabilities/total assets |
| | A17 | EBIT/interest expense |
| | A18 | Net operating cash flow/current liabilities |
| | A19 | Non-current liabilities/(non-current liabilities + owners' equity) |
| | A20 | Net operating cash flow/total liabilities |

| Category | # | Ratio definition |
|---|---|---|
| Operational capabilities | A21 | Main business income/average total assets |
| | A22 | Sales revenue/average fixed assets |
| | A23 | Main business cost/average inventory |
| | A24 | Main business income/average accounts receivable |
| | A25 | Sales revenue/average current assets |
| | A26 | Cost of sales/average accounts payable |
| | A27 | Sales revenue/average working capital |
| Capital expansion | A28 | Net increase in cash/number of ordinary shares |
| | A29 | Net assets/number of ordinary shares |
| | A30 | Net profit/number of ordinary shares |
| | A31 | Capital reserves/number of ordinary shares |
| Finance structure | A32 | Current assets/total assets |
| | A33 | Fixed assets/total assets |
| | A34 | Shareholders' equity/fixed assets |
| | A35 | Current liabilities/total liabilities |
| | A36 | Cash flow/total assets |
| | A37 | Accounts receivable/total liabilities |
| | A38 | Current liabilities/shareholders' equity |
| | A39 | Working capital/total assets |
| Trade credit | A40 | Accounts payable/total assets |
| | A41 | Accounts receivable/total assets |
| | A42 | Net trade credit (AP − AR)/total assets |

**C. Country & Macro Node: Static Attributes**
Country Names, Cities

**D. Country & Macro Node: Temporal Attributes**
- Event Information stored on edges (event count, relevance sum, sentiment mean, event classification).
- Disaster quarterly series stored on edges (event count, deaths, damage, day lasted, etc).
- Quarterly trade series stored on country nodes (exports, imports, total, net).

**Figure 2. Summary of Network Structure and Attributes**

Directed supplier and customer edges connect firms, ownership edges connect parents and subsidiaries, and headquarters edges connect firms to their home countries. Quarterly event, disaster, and trade summaries attach to country nodes, and edge weights reflect the relationship evidence intensity reported by the data provider.

The prediction targets are quarterly changes in revenue, inventory, and profitability rather than levels, since levels mainly reflect firm size [16] and changes remain comparable across firms. Changes also carry predictive information of their own, as momentum research shows for returns [27], and they directly measure the impact of a shock. A focal firm qualifies as a prediction target when it belongs to the collected company list, carries a complete 60-quarter record, and has at least two suppliers and two customers in the network, so that the graph layers have neighbor sets to attend over, and 116 focal firms meet these conditions. A sample observed through quarter t forecasts the change H quarters ahead for H = 1 and 2, and a target is valid only when both endpoints are observed. We split the 60 quarters into contiguous training, validation, and test windows of 48, 6, and 6 quarters, train on all firms, and evaluate on the 116 target firms. The six test quarters from 2023Q3 to 2024Q4 provide six forecast origins per firm and horizon, and pooling the three targets yields roughly 2,040 valid evaluation points per horizon, around 680 per target.

### 4.3. Proposed Model

The proposed model consists of three major components, namely (1) multi-source encoding and gated fusion, (2) typed graph propagation, and (3) patch-based delta forecasting (Figure 3).

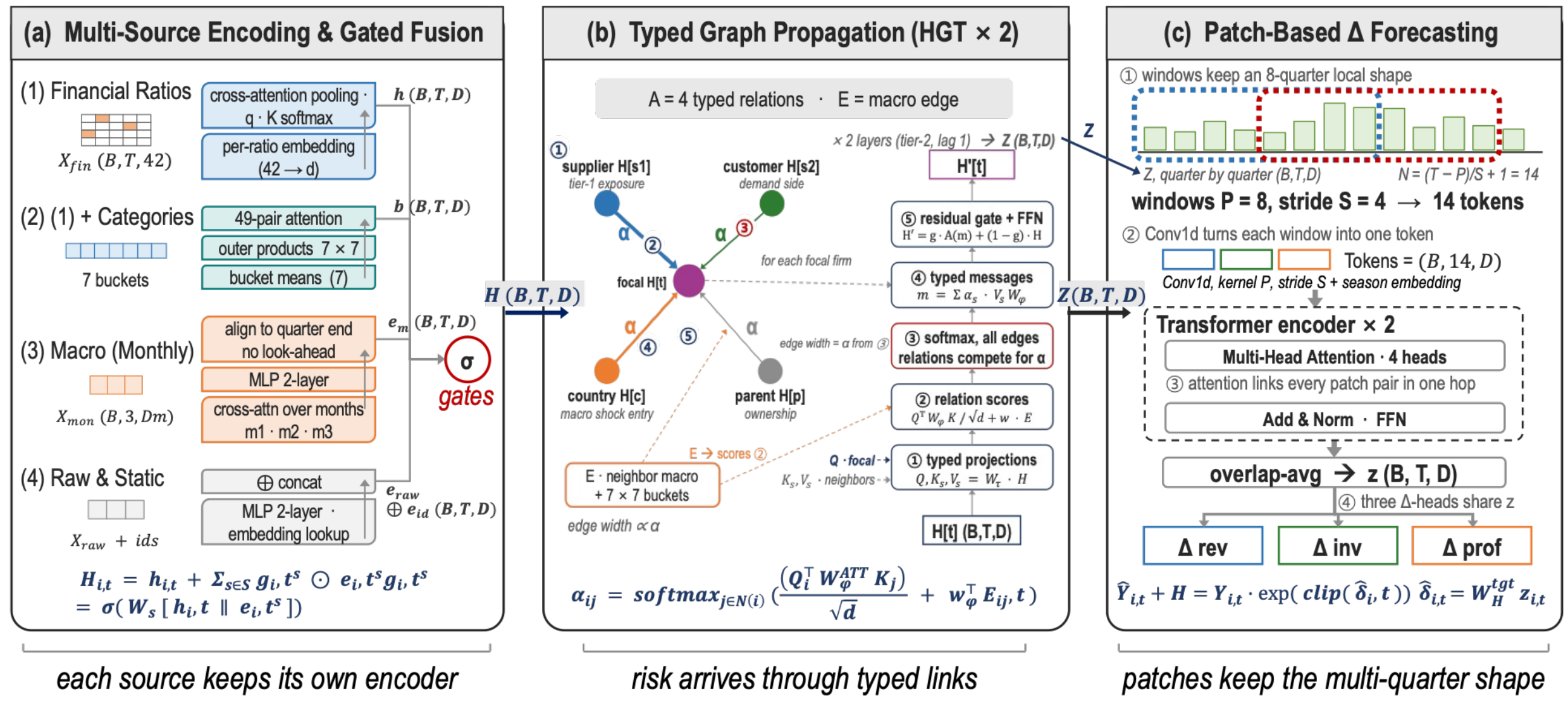

**Figure 3. Overall Architecture of the Proposed Model**

Formally, given the network G and the multi-source inputs observed through quarter t, the model outputs the changes in revenue, inventory, and profitability at horizons H = 1 and 2. The first component assembles a quarterly state for each firm from its own reports and its environment, the second revises this state with information from connected firms, and the third maps the revised histories to quarterly changes, mirroring the propagation process documented in our review, in which shocks form at individual firms, travel along typed relationships, and appear in fundamentals with a delay. We adopt the attention parameterization of HGT [25] and the patch tokenization of PatchTST [40], and our contribution lies in the gated multi-source fusion, the macro edge features, and their integration into one pipeline.

**4.3.1. Multi-Source Encoding and Gated Fusion.** Because the data sources differ in frequency, dimensionality, and missingness, we encode each source separately before any fusion. A ratio encoder embeds each quarterly ratio and pools the embeddings with category-aware attention, weighted by the observation masks so that unreported ratios are excluded rather than imputed. The mask treatment follows Che et al. [8], and the attentive pooling over ratio categories adapts Che et al. [7]. A raw-value encoder embeds the dollar levels of revenue, cost, and inventory, which ratios alone would hide. A bucket encoder applies attention over the 49 pairwise products of the seven ratio-category means to capture joint conditions such as simultaneous deterioration in solvency and profitability. A macro encoder attends from the quarterly state to the three monthly macro vectors of the quarter, so intra-quarter movements enter without look-ahead. We then fuse the encodings into one quarterly firm state through learned gates

$$H_{i,t} = h_{i,t} + \sum_{s \in S} g_{i,t}^s \odot e_{i,t}^s \,, \quad g_{i,t}^s = \sigma\left(W_s \left[h_{i,t} \parallel e_{i,t}^s\right]\right)$$

where h is the ratio encoding of firm i in quarter t, $e^s$ is the encoding of auxiliary source s, S = {raw, bucket, macro} collects the auxiliary sources, σ is the sigmoid function, ‖ denotes concatenation, and ⊙ denotes elementwise scaling. Because each gate is computed per source and quarter, the fusion can emphasize macro signals in shock quarters and firm fundamentals in stable ones. This gating extends prior multisource fusion designs, which enrich inputs but weight them uniformly [48, 58]. Without the gates, missing sources would enter as zeros and contaminate calm quarters.

**4.3.2. Typed Graph Propagation.** To propagate risk along relations whose effects differ by type and direction, we apply two heterogeneous graph transformer layers to the fused states [25]. For a focal firm i and a neighbor j connected by relation φ, node-type-specific projections generate the query Q, key K, and value V, and each edge receives a relation-specific score augmented by macro edge features, implementing typed, direction-specific propagation

$$\alpha_{ij} = softmax_{j \in N(i)} \left( \frac{Q_i^\top W_\varphi^{ATT} K_j}{\sqrt{d}} + w_\varphi^\top E_{ij,t} \right)$$

where N(i) collects neighbors of every type, d is the query and key dimension, E carries neighbor-country macro conditions and cross-firm bucket interactions, and W and w are the attention matrix and edge-feature weights of relation φ. Because one softmax normalizes the scores of every neighbor and relation jointly, supplier, customer, ownership, and headquarters edges compete for attention, and each relation learns its own direction-specific transmission strength. This addresses the uniform aggregation limitation from our review and reflects the asymmetric transmission documented in finance [12, 22]. The attention-weighted messages then update each focal state through a gated residual

$$H'_{i,t} = FFN\left( g \odot A\left( \sum_{j \in N(i)} \alpha_{ij} V_j W_\varphi^{MSG} \right) + (1 - g) \odot H_{i,t} \right)$$

where A(·) is a type-specific output projection, FFN is a position-wise feed-forward network, and the gate g is initialized near zero, so network context supplements rather than replaces firm fundamentals. The first layer aggregates tier-one partners within quarter t, and the second layer repeats the computation with the first layer's outputs at $t - 1$ as its keys and values, so a tier-two shock reaches the focal state with a one-quarter lag. We restrict the depth to two layers, matching the documented reach of sub-tier effects and limiting over-smoothing [14, 51]. Without typed edges, supplier and customer shocks would blend into one averaged signal.

**4.3.3. Patch-Based Delta Forecasting.** Because each firm contributes at most 60 quarterly observations, a patch tokenizer groups the graph-updated sequence into patches of eight consecutive quarters with stride four, which yields roughly 14 overlapping tokens per firm, preserves the local shape of multi-year patterns, and shortens the sequence entering attention fourfold [40]. A two-layer transformer encoder with four attention heads models the dependencies among tokens, and a seasonal embedding marks each token's calendar quarter. The overlapping patches are then averaged back into one representation z per firm and quarter, and a linear head maps z to the predicted changes in revenue, inventory, and profitability, with a separate model trained for each horizon

$$\hat{Y}_{i,t+H} = Y_{i,t} \cdot exp\left( clip\left(\hat{\delta}_{i,t}\right) \right), \quad \hat{\delta}_{i,t} = W_H^{tgt} z_{i,t}$$

where Y is the last observed level and clip bounds extreme updates. Dollar targets are predicted as clipped log changes, which keeps forecasts positive and comparable across firm sizes, and the profitability ratio as a bounded additive change. Without patching, each quarter would enter attention alone and the encoder would memorize noise rather than shape.

All components are trained end to end by minimizing a weighted MAE of the predicted changes, which resists the extreme quarters that dominate squared losses [25], with a small penalty that shrinks predictions toward the naive no-change forecast, using AdamW and validation-based early stopping.

## 4.4. Evaluation Design

We evaluate the proposed model against benchmark methods, summarized in Table 2.

| Family | Models | Purpose | References |
|---|---|---|---|
| ML baselines | LR, MLP | Performance floor | Standard baselines |
| Temporal | GRU, LSTM, Transformer, TCN, PatchTST | Per-firm sequence modeling | [11], [24], [46], [2], [40] |
| Graph | GCN, GraphSAGE | Static message passing | [31], [20] |
| Hybrid | SAGE + GRU / LSTM / Transformer / PatchTST | Propagation plus temporal encoder | [20], [40] |
| Proposed | HGT + PatchTST | Typed two-hop propagation plus patch encoding | [25], [40] |

**Table 2. Benchmark Methods by Family**

Machine learning baselines, namely linear regression (LR) and a multilayer perceptron (MLP), establish the performance floor. Temporal models, including the temporal convolutional network (TCN), isolate sequence modeling, graph models isolate message passing, and hybrid models add propagation to temporal encoders. All models follow the Section 4.2 protocol. We report MAE in millions of dollars for revenue and inventory and the symmetric mean absolute percentage error (sMAPE) for profitability, because absolute errors of a ratio are not comparable across firms. The metrics are

$$MAE = \frac{1}{n}\sum_{k=1}^{n} |\hat{Y}_k - Y_k| \, , \quad sMAPE = \frac{1}{n}\sum_{k=1}^{n} \frac{2|\hat{Y}_k - Y_k|}{|Y_k| + |\hat{Y}_k|}$$

where n counts the evaluation points of a target and horizon and $\hat{Y}$ and Y denote the forecast and the realized value. This sMAPE variant keeps absolute values in the denominator, so it is bounded by two and stays defined when margins approach zero [26]. We report single-run point estimates and defer formal significance tests to the full study.

## V. Experiments and Results

We present the results in three comparisons, namely the temporal baselines, the graph and hybrid baselines, and the composition of errors across targets and horizons. The sequence reverses the order of our review, because the temporal contrast isolates the tokenization choice first, and the graph contrast then adds propagation on top of it. Table 3 reports all benchmark families at both horizons, with the best value per column in bold.

| Family | Model | Rev. H1 | Inv. H1 | Prof. H1 | Rev. H2 | Inv. H2 | Prof. H2 |
|---|---|---|---|---|---|---|---|
| ML baselines | LR | 544.8 | 222.3 | 0.149 | 692.2 | 266.8 | 0.184 |
| | MLP | 547.0 | 224.5 | 0.147 | 698.4 | 269.4 | 0.186 |
| Temporal | GRU | 383.0 | 172.9 | 0.148 | 521.2 | 232.3 | 0.187 |
| | LSTM | 390.5 | 176.5 | 0.153 | 510.9 | 227.9 | 0.192 |
| | Transformer | 424.3 | 177.7 | 0.149 | 712.8 | 271.8 | 0.204 |
| | TCN | 376.7 | 179.6 | 0.147 | 492.1 | 227.0 | 0.182 |
| | PatchTST | 227.5 | 174.6 | 0.054 | 313.1 | 218.3 | 0.084 |
| Graph | GCN | 482.9 | 186.5 | 0.147 | 623.4 | 247.8 | 0.176 |
| | GraphSAGE | 468.5 | 195.0 | 0.144 | 574.3 | 242.0 | 0.183 |
| Hybrid | SAGE + GRU | 377.8 | 177.0 | 0.154 | 525.7 | 235.2 | 0.185 |
| | SAGE + LSTM | 363.3 | 172.4 | 0.145 | 502.1 | 230.6 | 0.187 |
| | SAGE + Transformer | 413.6 | 176.3 | 0.145 | 712.8 | 271.8 | 0.204 |
| | SAGE + PatchTST | 246.2 | 176.4 | 0.066 | 295.2 | 236.0 | 0.084 |
| **Proposed** | **HGT + PatchTST** | **173.8** | **162.8** | **0.047** | **282.1** | **201.5** | **0.069** |

**Table 3. Forecast Accuracy at One- and Two-Quarter Horizons**

*Note. Rev. and Inv. report MAE in $M, Prof. reports sMAPE, and H is the forecast horizon in quarters. Bold marks the best value per column.*

The proposed model achieves the lowest error on every target at both horizons. At the one-quarter horizon, it records a revenue MAE of $173.8M, an inventory MAE of $162.8M, and a profitability sMAPE of 0.047, against $227.5M, $172.4M, and 0.054 for the strongest baselines on each target.

Recall that our review faulted temporal models for treating each firm in isolation and each quarter as one token. Against these benchmarks, the proposed model reduces revenue MAE by 54.6 percent relative to the best recurrent model ($173.8M against $383.0M for GRU) and by 23.6 percent relative to PatchTST, the strongest temporal baseline. The profitability results isolate the tokenization choice. Every point-wise model, including the plain Transformer, records an sMAPE of 0.14 or higher, while every patch-based variant stays below 0.09, indicating that quarterly margins move in multi-quarter patterns that point-wise tokens smooth away. Whether the remaining margin comes from the network is the next comparison.

Recall also that our review faulted standard GNNs for uniform neighbor weights. Against the graph benchmarks the advantage is larger still. GCN and GraphSAGE record revenue MAEs of $482.9M and $468.5M at one quarter, trailing even the simple GRU, so propagating a static snapshot without temporal encoding adds little. Pairing the same GraphSAGE module with the patch encoder recovers much of the gap ($246.2M), and replacing its uniform pooling with typed, direction-specific propagation yields a further 29.4 percent reduction to $173.8M and a profitability improvement from 0.066 to 0.047. Because this hybrid shares the temporal encoder with the proposed model, the remaining margin isolates typed, direction-specific propagation and supports our argument that supplier and customer relations should not share aggregation weights.

Improvements are smallest for inventory, at 5.6 percent over the best baseline at the one-quarter horizon and 7.7 percent at the two-quarter horizon. Inventory is the noisiest target, since reported levels mix order batching, write-downs, and management policy only weakly related to network conditions, so every model faces a similar floor. The horizon comparison shows the opposite pattern for the network. As firm-internal momentum fades, the revenue margin over the strongest baseline narrows from 23.6 percent, over PatchTST at one quarter, to 4.4 percent, over the SAGE + PatchTST hybrid at two quarters, while the profitability margin widens from 13.0 to 17.9 percent, consistent with the one-quarter lag built into the second propagation layer. Taken together, typed, direction-specific propagation and patch-based encoding address the limitations that motivated them. With respect to RQ2, heterogeneous multi-frequency signals can be integrated in one graph-based framework, and the integration yields measurable gains over strong firm-level baselines.

## VI. Conclusion and Future Directions

Supply chain stress reaches the reported financials of connected semiconductor firms with a delay, and the current AI cycle is writing that pattern across foundries, memory, and infrastructure. Prior research offers no direct forecast of this impact. Detection studies stop before financial consequences, financial risk models target market-based labels rather than operating outcomes, and existing methods capture network structure or short quarterly histories, but not both.

We addressed these gaps with a heterogeneous graph patch transformer evaluated on 116 focal semiconductor firms in a 15,186-company network. The model achieved the lowest error on every target at both horizons, cut revenue MAE by 23.6 percent against the strongest baseline at one quarter, and widened its profitability margin from 13.0 to 17.9 percent at two quarters, where analyst coverage thins and foresight matters most. These gains concentrate exactly where the design predicts, in typed, direction-specific propagation over patch-encoded histories, so the study contributes both a multi-tier testbed that links network structure to reported fundamentals and early evidence that supply chain risk is forecastable in financial terms.

These results carry practical implications, which a few examples make concrete. Semiconductor capacity is committed years in advance, so a forecast one to two quarters ahead cannot create supply but buys time to prepare. When a forecast warns that a critical supplier will weaken as demand rises, the firm can extend trade credit or begin qualifying an alternative source while both responses are still cheap. Because advanced foundries, memory, and equipment concentrate in a handful of firms, the same forecasts trace exposure past the industry's leaders to the supporting firms around them, beyond the tier one where most companies' visibility ends. That map matters to governments weighing geopolitical exposure as much as to the firms inside it. Investors have a parallel use for the same signal. Prices incorporate supply chain links slowly, so a forecast of fundamentals can run ahead of that adjustment in technology-led markets. None of these uses follows from architecture alone. The gains arose because each component answers an industry characteristic, typed relations for a multi-tier value chain, patch encoding for short quarterly histories, and a different industry would demand a different composition.

We will extend the testbed through 2026Q2 and re-estimate all benchmarks, test how proximity to shocked nodes and shared sub-tier dependencies shape a firm's response against realized disruptions (RQ1), and attach exposure explanations through the learned gates and attention weights. Extending the network to deeper tiers and benchmarking against analysts' forecasts at longer horizons complete the agenda. Reading the network early, rather than after stress appears in reported financials, is the capability this study begins to build.